# GENERATIVE TRANSFORMERS FOR DESIGN CONCEPT GENERATION

**Qihao Zhu**
Data-Driven Innovation Lab
Singapore University of Technology and Design
qihao_zhu@mymail.sutd.edu.sg

**Jianxi Luo**
Data-Driven Innovation Lab
Singapore University of Technology and Design
jianxi_luo@sutd.edu.sg

November 7, 2022

## ABSTRACT

Generating novel and useful concepts is essential during the early design stage to explore a large variety of design opportunities, which usually requires advanced design thinking ability and a wide range of knowledge from designers. Growing works on computer-aided tools have explored the retrieval of knowledge and heuristics from design data. However, they only provide stimuli to inspire designers from limited aspects. This study explores the recent advance of the natural language generation (NLG) technique in the artificial intelligence (AI) field to automate the early-stage design concept generation. Specifically, a novel approach utilizing the generative pre-trained transformer (GPT) is proposed to leverage the knowledge and reasoning from textual data and transform them into new concepts in understandable language. Three concept generation tasks are defined to leverage different knowledge and reasoning: domain knowledge synthesis, problem-driven synthesis, and analogy-driven synthesis. The experiments with both human and data-driven evaluation show good performance in generating novel and useful concepts.

## 1 Introduction

At the early stage of design, the quantity and diversity of design concepts are essential for designers to explore new design opportunities departing away from existing designs [1]. However, the well-recognized features in existing designs can fixate designers' thinking and limit their capability of generating novel design concepts [2-4]. Traditional approaches to overcome design fixation include brainstorming, mind mapping, and design heuristics [5-8] which aim to improve creative thinking and help designers to think out-of-the-box. Meanwhile, the limitation of designers' knowledge base is also an important source of design fixation [9-10]. Figure 1 illustrates a highly summarized view of the human cognitive process of design concept generation, in which prior knowledge is transformed into novel concepts through creative reasoning. Therefore, methods and tools that could augment the knowledge and creative reasoning aspects of the process may enhance design concept generation.

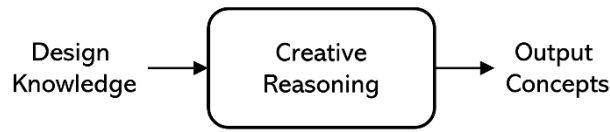

Figure 1: Design concept generation process

In the era of data-driven innovation [11], researchers have explored various data-driven approaches and data sources to augment design for innovation. For instance, technical or conceptual data have been mined to uncover designers' creative thinking heuristics during the ideation stage [7-8, 12-14]. Various approaches have been proposed to retrieve knowledge from data that provoke designers for creative and cross-domain design concept generation [9-10, 15-16]. Other works explored generative approaches of function synthesis based on design repositories [17-24] or deep generative models based on product images or geometry data [25-31]. To date, few studies have leveraged the creative reasoning and prior knowledge from data simultaneously and transformed them into understandable design concepts. Thus, the inspiration that designers can draw using these methods is limited.

Herein, we present a data-driven design concept generation method. Our method utilizes the pre-trained language model (PLM), a cutting-edge technology in artificial intelligence (AI). We show that PLM can learn the knowledge and creative reasoning from the textual data in design repositories and automatically generate novel concepts in natural language. Three sets of design concept generation experiments will be presented to showcase the performance of the method on that basis. These experiments also demonstrate the AI-based generative conceptual design process, i.e., how designers can collaborate with AI to explore potential solutions and develop design concepts.

This paper is organized as follow. In Section 2, we review the existing design concept generation research. Section 3 will introduce the natural language generation technology that we use in the paper. Section 4 presents our method in detail, followed by the experiment settings and results in Section 5. Finally, Section 6 gives a summary discussion and Section 7 discusses the limitation and the potential future improvement to the method.

## 2 Literature Review

The computer aids for design concept generation can be categorized into three types according to their potential roles in the human-computer collaborative relationship. The roles of methods or tools are a) to guide, b) to stimulate, and c) to generate. Figure 2 shows different modes of computer-aided conceptual design in terms of human-computer collaboration. Figure 2 (a) is the most common scenario where human designers gather to generate ideas, and record, analyze and improve them with the aid or guide of computers. In scenario (b), computers provide stimuli to inspire human designers to generate ideas. Existing knowledge- or heuristics-based computer tools can facilitate such a process. In scenario (c), a computer agent (i.e., a virtual creative designer) generates concepts for human designers to evaluate, select and improve. The output could also provoke designers to be more creative with the extended knowledge and thinking modes. Scenario (c) is an expected application of the method in this paper, in which the virtual designer works as a brainstorming collaborator generating new design concepts at a quality and quantity far beyond human capacity.

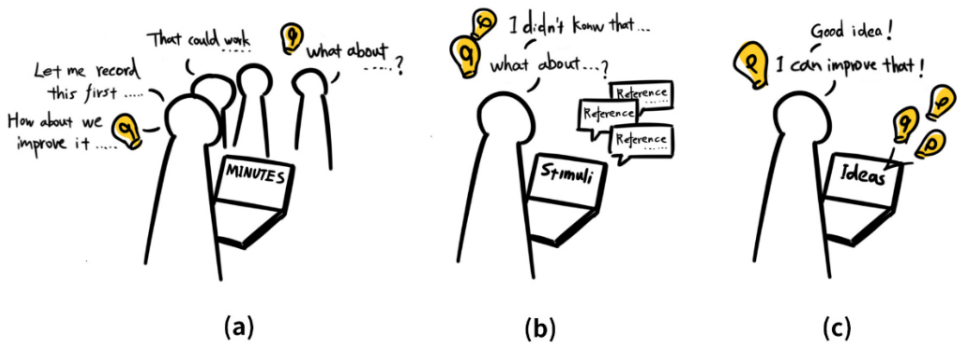

Figure 2: Human-computer collaboration in different ideation approaches: (a) Brainstorming (b) Stimuli-based approach (c) Generative approach

## 2.1 Guiding Design Concept Generation

The methods to guide are instructively involved in the generation of design concepts, providing design rules or guidelines to the activities of the designers. The common group-based innovation approaches of brainstorming, C-sketch, and mind mapping fall into this category [5-6, 32]. Although these methods are considered less efficient in the data-driven era, their variants and computer-aided improvements have been proposed, e.g., data-driven brainstorming [33]. Bonnardel and Didier [34] proposed two variants of brainstorming, encouraging designers to focus on the evocation of both the design ideas and the constraints related to the design problem. Wang et al [35] introduced a creative agent that generates feedback and recommendations for designers during brainstorming. Lee et al. [36] developed Speech2Mindmap that automatically transforms speech data collected from brainstorming sessions into mind maps. Empirical studies have shown great potential of AI tools to guide a more explorative design strategy [37-38] and manage team-based design process efficiently [39]. However, these tools can only guide human to be more creative without providing any creative reasoning themselves. The creative thinking capability and knowledge input of the human designer group are still essential in the early-stage design process.

## 2.2 Stimulating Design Concept Generation

The methods to stimulate the concept generation process provide inspirational stimuli to provoke designers to conceive new concepts. These include design heuristics that provide critical creative thinking and knowledge pieces that extend the knowledge scope of designers. For design heuristics, Altshuller [8] proposed the 40 TRIZ principles based on patent innovations that focus on solving the technical contradictions in engineering design. Yilmaz et al. [13] derived 77 design heuristics from a dataset of 3,457 award-winning products and concepts designed by experts. Such heuristics aim to discover the patterns of how experienced designers design conceptual products. Jin and Dong [14] extracted 10 design heuristics as stimuli from RedDot award-winning design concepts to help digital designers overcome design fixation. Moreover, [40] used the heuristic procedure of morphological matrix to guide designers to generate a variety of design options. The heuristic includes identifying and combining alternative solutions to multiple functions of a product. Stone et al. [41] also propose heuristic methods using functional modelling derived from functional basis and function chains.

Knowledge-based stimulation approaches have been focusing on the retrieval and mapping of source knowledge into the target design domain [42]. He et al. [43] tested the use of word clouds as stimulators to inspire ideation. Luo et al. [9-10] introduced a computer-aided ideation tool InnoGPS to guide the provision of design stimuli. InnoGPS enables designers to search for inspiration in the patent database by the knowledge distance to the design problem or interest. Sarica et al. [15-16] developed a technology semantic network for identifying technical white space according to the semantic distance between the design target and stimulus. Recently, several large semantic networks and engineering knowledge graphs have been created as knowledge infrastructures to aid design representation, reasoning, and stimulation [44-46].

Most stimuli-based research focuses on either knowledge retrieval as the source of inspiration or creative thinking heuristics in concept generation. However, there are contexts where both sides should be considered together to perform successful conceptual design. Especially, in biologically inspired design research, to bridge the gap between the domains of biology and engineering will require both the knowledge of biological systems and the methods to transfer them [47-48]. For example, Vattam et al. [49] introduced DANE, which represents the knowledge of natural biological systems in text descriptions and images. In addition, to capture the functions of the biological system, structure and behavior models are added. AskNature is a web-based tool that offers not only a wide variety of scientific knowledge, but also biomimicry strategies to help designers work on bio-inspired design [50]. Other works also suggest that visual stimuli like images and sketches can affect design performance from the initial problem-solving idea generation to the visual embodiment design of the product [51-54].

## 2.3 Automatically Generating Design Concept

As a result of limited digital knowledge and computing algorithms for utilizing in this process, fully automated approaches are considered challenging during the conceptual design phase [17, 55]. This could result in a stimuli-based approach that is still dependent on humans to infer from the initial information provided by computers. Existing attempts exploring generative computing approaches mainly include the creation of function structure and the optimization of well-established concepts [17]. For example, computer-aided functional analysis utilizes a design repository including component connections and functionality of the recorded products [17-19]. Function structures can be generated based on the knowledge and relations stored in the repository. The resulted function structure can either be represented in a function model [20-22] or a function-means tree [23-24]. This approach has a significant advantage of leveraging the functional knowledge and structural relations from the design data. It enables designers to quickly come up with new concepts.

However, the reported design repository utilized in this kind of research usually contains no more than a few hundred domain-focused products [20, 21, 22]. This is probably due to the amount of work involved in constructing such a function-based design repository. Thus, the approach of function-based synthesis for automatic concept generation can face two major drawbacks: First, the output function structure is commonly represented in abstract diagram of functions and components and well-trained engineers are needed to make sense of the concepts. Today, collaborative design scenarios involve members of diverse expertise working together through negotiation and evaluation [56]. Thus, the concepts need to be presented in a more understandable way for different expertise to understand. Secondly, the size and scope of the limited human-curated design repository data can prevent designers from exploring the full range of cross-domain design options.

Automatic concept generation may also take the form of optimizing well-established concepts to achieve predetermined objectives. This includes topology optimization and generative shape synthesis [31, 57]. Topology optimization aims to optimize the distribution of the material in a given design space according to a certain objective of product performance, e.g., reduce weight, reduce stress, and improve stiffness [58]. This also requires designers to set up the design space with constraints for optimization. Shape synthesis is traditionally implemented through shape grammar [59-60] or parametric approaches [61-62]. Both approaches are based on computational algorithms to transform human input of objects, rules, or parameters into the desired 2D or 3D geometry. The recent advance in artificial intelligence (AI) has led to deep generative models like Generative Adversarial Network (GAN) [63] and Variational Autoencoder (VAE) [64], which have received increasing interest from the engineering design community.

Research efforts utilizing these novel techniques have been seen in topology optimization and generative shape synthesis [31]. Oh et al. [25] and Nie et al. [26] integrate GAN into topology optimization and result in a significantly decreasing computational cost compared to traditional approaches. Burnap et al. [27] introduced VAE to the generation of new concepts of 2D vehicle profile design. Yilmaz and German [28] used GAN to conditionally generate the design of the airfoil section. Li et al. [29] developed a recursive autoencoder that encodes and synthesizes 3D shapes. Using an iterative training approach, Shu et al. [30] optimize the 3D shape of aircraft designs with GANs and Computational Fluid Dynamics, which take the generated results with high performance to update the dataset for the next loop of training. These methods utilizing deep generative models can learn from the geometric training data and synthesize visual representations of design concepts. They can be potentially applicable to the optimization and customization of many design domains [31]. Readers are referred to [31] for a detailed review of deep generative models in engineering design. However, a major limitation of these models is that they are currently only capable of learning and generating spatially visualized representations of design concepts, e.g., images, point clouds, and meshes. These representations are mainly useful for embodiment and detail design and may cause design fixation if applied in earlier stages [4].

Figure 3 summarizes the concept generation methods and tools according to their roles in human-computer collaboration, the source of knowledge and reasoning input, and the form of the computational output if applicable. In Figure 3, 'knowledge' refers to both the internal and external knowledge space that is potentially applicable to the target design space, while 'reasoning' refers to any kind of mindset, guideline, heuristic thinking, or computational algorithm that transforms the knowledge into novel design concepts. Our method is located at the top right of Figure 3 and focused on the verbal generation approach using the cutting-edge language models.

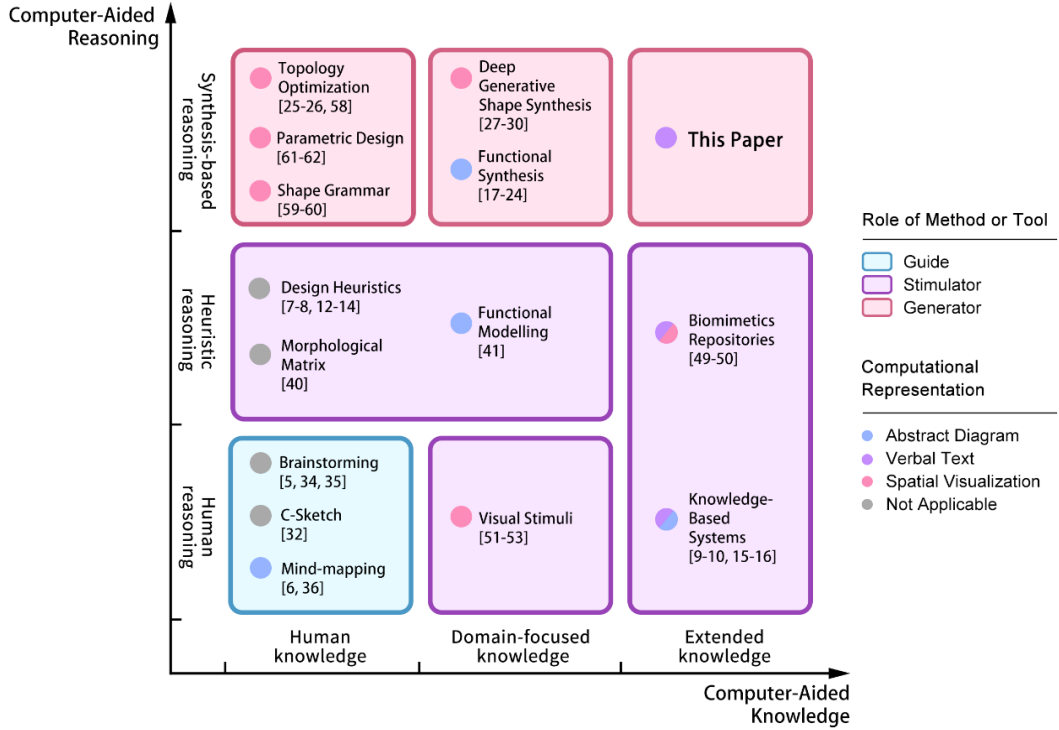

Figure 3: Taxonomy of concept generation methods and tools

Our work aims to contribute to natural language processing (NLP) in design research [65], with a focus on generative transformers and natural language generation (NLG) for design concept generation. Particularly, we experiment with generative pre-trained transformers (GPT) to learn the design knowledge and reasoning from task-oriented datasets, and then generate high fidelity design concept descriptions in natural language.

## 3 Natural Language Generation

In this section, we briefly review the natural language generation (NLG) technology we use in this research as well as its evaluation metrics. Especially, the popular series of generative pre-trained transformers from OpenAI, including GPT-2 and GPT-3, are introduced, as they will be applied in the experiments in this paper.

### 3.1 Text-to-Text and Data-to-Text NLG

NLG is considered as a computer program that generates natural language as output [66]. According to [66], depending on the input, there are two major instances of NLG: text-to-text and data-to-text generation. Text-to-text is what takes in the existing text as input and then outputs new pieces of text. Examples of text-to-text generation include machine translation (e.g., [67]), text summarization or simplification (e.g., [68]), paraphrasing (e.g., [69]), and so on. On the other hand, data-to-text generation takes in non-linguistic data as input. Applications of data-to-text have been seen in varied fields, e.g., generating news based on election data [70], and generating personalized suggestions based on user input and web data [71]. Considering real-world applications, data-to-text generation has been a more popularly recognized type of NLG. Accordingly, some researchers define NLG as only the generation of data from non-textual input [72].

## 3.2 Pre-Trained Language Model

Pre-Trained Language models (PLMs) are language models that have been trained with a large textual dataset collected from varied sources. The training data could include but not limited to Wiki, books, and web data. The pre-trained model can be applied to specific language-related tasks [73]. In recent years, transformer-based language models have been achieving state-of-the-art performance on many tasks. Originally proposed by [74], transformer is the most advanced neural network architecture for natural language processing (NLP) and has quickly become the dominant for NLG [75]. By contrast to the recurrent neural network (RNN) and the long short-term memory (LSTM), two of the most popular neural network architectures until recently, transformer accounts for vanishing gradients [76]. Vanishing gradients were perceived as causing context loss when processing longer texts in RNN and LSTM.

Moreover, the transformer enables parallel training. With the training data and model architecture becoming larger in size, it can capture longer sequence features and therefore result in much more comprehensive language understanding and generation [77]. The state-of-the-art network architecture of transformers, as well as the huge pre-training datasets, offer PLMs with not only the capability of learning human language, but also the knowledge and logic that come with it. By applying PLMs, NLG systems could leverage and transform a wide range of knowledge without the need for manual formulation and inference.

Another significant feature of transformer for NLG is that it blurs the boundary between data-to-text and text-to-text NLG models. For instance, Radford et al. [78] shows outstanding performance of the GPT-2 transformer model from OpenAI for text-to-text generation tasks such as translation and summarization. Peng et al. [79] shows the model after fine-tuning is also capable of performing data-to-text dialogue generation tasks.

Although transformer-based PLM is a rather new technique in NLG, some applications have already been seen in different fields. Amin-Nejad et al. [80] use transformer models to create structured patient information to augment medical datasets. Lee and Hsiang [81] use GPT-2 with fine-tuning to generate patent claims. Fang [82] also uses GPT to generate ideas for content creators. However, according to [31], the application of transformers is still a wide-open space for engineering design tasks. Our work fills this gap.

## 3.3 Generative Pre-trained Transformer

Now we introduce the most popular series of transformers in NLG tasks: the generative pre-trained transformer [77-78, 83], or GPT for short. Later in this paper, we apply and compare the 2nd and 3rd generations of GPT, i.e., GPT-2 and GPT-3, in design concept generation tasks.

GPT-2 uses the two-step training strategy of pre-training and fine-tuning, following [84]. The workflow is shown in Figure 4(a). During the pre-training step, the model is trained on a massive text dataset called WebText, which is collected from millions of web pages [78]. This results in a comprehensive language model that can perform general language completion tasks. According to OpenAI, the largest GPT-2 pre-trained model has 1.5 billion parameters. For downstream NLP tasks, the pre-trained model then needs to be fine-tuned given a customized and task-oriented dataset. The fine-tuned model is trained through repeated gradient updates using a large dataset of a corpus of the example task. This process updates the weights of the pre-trained model and stores them for the use of the target task. However, the large dataset customized for the target NLP task may be unavailable or difficult to collect.

GPT-3 is trained on a mixture of datasets containing 400 billion tokens and has a maximum of 175 billion parameters, over a hundred times larger than GPT-2. Compared to its precursor, GPT-3 can perform a wide range of NLP tasks without the need for fine-tuning. GPT-3 is capable of zero-shot learning, one-shot learning, and few-show learning [77]. In zero-shot learning, the model is given only the description of the task and generates the answer in natural language. In one-shot learning, the model is given a single example of the task in addition to the description. In few-shot learning, the model learns from multiple examples before performing the task. In all three modes, no gradient updates are performed [77]. The training and generation process of GPT-3 is shown in Figure 4(b), where the prompts are the task descriptions and examples given to the model for zero-shot, one-shot, or few-shot learning.

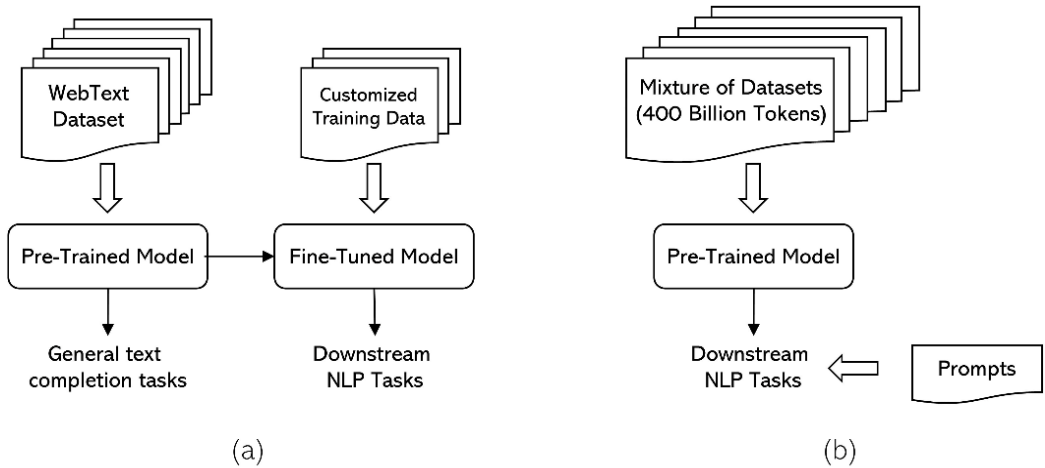

Figure 4: Training process of GPT-2 (a) and GPT-3 (b)

For text generation, the implementation of GPT requires several control parameters (Table 1) and can be categorized into three types by the aimed property of the generated text. The 'max_tokens' (also called 'length'), 'prompt' (also called 'profix'), and 'stop' (also called 'truncate') control the content and format of the text. These are essential parameters when customizing the task and defining the input of the NLG model. When conditionally sampling with a pre-determined prompt, the model will learn to set up the context and output results based on it. The 'temperature', 'top-k', and 'top-p' parameters control the randomness of the generated text. Higher randomness will result in more varied outputs. Finally, the 'presence_penalty' and 'frequency_penalty' parameters are only applicable for GPT-3 and control generation repetitiveness. By encouraging original topics and discouraging existing ones, the generated texts are more likely to deviate from the example and represent novel concepts.

Table 1: Main parameters of text generation with GPT

| Parameter | Explanation | GPT-2 | GPT-3 |
|---|---|---|---|
| max_tokens | The maximum number of tokens to generate in the completion | √ | √ |
| prompt | The prompt(s) to generate completions for | √ | √ |
| stop | Up to 4 sequences where the API will stop generating further tokens. | √ | √ |
| temperature | Higher values mean the model will take more risks | √ | √ |
| top_k | Random sampling the next token from the k most likely candidates [85] | √ | |
| top_p | The model considers the results of the tokens with top_p probability mass | √ | √ |
| presence_penalty | Positive values penalize new tokens based on whether they appear in the text so far, increasing the model's likelihood to talk about new topics | | √ |
| frequency_penalty | Positive values penalize new tokens based on their existing frequency in the text so far, decreasing the model's likelihood to repeat the same line verbatim | | √ |

*Explanations of parameters except top-k are from OpenAI[1]

Another well-known transformer-based PLM is the Bidirectional Encoder Representations from Transformers (BERT) [86], which was trained with Wiki and books data that contains over 3.3 billion tokens. BERT is commonly used in natural language understanding (NLU) tasks such as text classification and keyword extraction. However, as a masked language model, BERT is generally weak at NLG, because it can only learn the contextual representation of words [73].

### 3.4 NLG Evaluation

To facilitate NLG experiments and development, researchers have developed automatic evaluation metrics for rapid iteration of NLG techniques. A common approach is to measure the n-gram matching to represent the similarity between the machine-generated text and the human-written ground truth text. BLEU is a weighted geometric mean of n-gram precision scores [87]. ROUGE measure n-gram recall instead of precision [88]. METEOR is another n-gram based metric that calculates the harmonic mean of precision and recall [89]. Although the n-gram overlap approach is one of the most utilized metrics in NLG research, it only considers the surface level similarity between the reference and the generated text, i.e., are the words in the two paragraphs match or not?

More recent methods consider semantic level similarity by measuring text embeddings. Word Mover's Distance (WMD) is a measurement of the distance between the probability distribution of two vector spaces extracted from the word embeddings of documents [90]. Sentence Mover's Distance (SMD), similar to WMD, also measures the embeddings of two documents, but it is based on the embedding of sentences instead of words [91]. The embedding-based approaches go beyond surface-level measurement and provide a more intuitive scale of how two documents are related to each other. However, these methods still require ground truth references to compare the generated results. Ground truth is not always available in real-world applications because there is no preferred answer for many text generation tasks (e.g., open-ended question answering). Other works explored machine-learned metrics (e.g., [92-93]) to evaluate open-ended text generation without the need for baseline references, but they require human-annotated samples to train or fine-tune such evaluation models.

## 4 Method

### 4.1 Concept Generation

In this paper, we experiment with the applications of different GPT models in different design concept generation tasks. GPTs are trained to pick up patterns in human language and such patterns will also be reflected in the generated text [94]. Through properly preparing the fine-tuning data or prompts, we can force the models to connect remote concepts that rarely appear together in existing knowledge and synthesize them into novel new design concepts.

---
[1] https://beta.openai.com/docs/api-reference/

In psychology and creative cognition literatures, there have been many contributions on creative thinking that involves connecting concepts that are weakly related, both theoretically (e.g., [95]) and empirically (e.g., [96]). This pattern has also been discovered in the research of novel design. Simonton [97] suggests that the unconventional combinations of prior technologies could results in novel innovations. He & Luo [98] statistically found the most valuable inventions tend to embed extremely novel combinations of prior knowledge. Luo & Wood [99] also revealed a trend that patented inventions have been combining the knowledge of broader domains over the past three decades. Thus, by teaching GPTs to connect remote knowledge and generate solutions diverse from existing ones, creative reasoning for novel concept generation can be supported. In fact, a recent study empirically suggests that GPT-3 is capable of divergent thinking to perform creative tasks [100]. Therefore, by customizing the dataset for fine-tuning or few-shot learning and the prompts, we control the learning and reasoning of GPT to generate novel concepts in varied design tasks. Figure 5 depicts the general framework of our experiments.

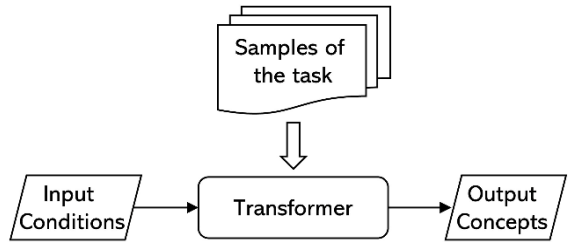

Figure 5: Experimental Framework

Three essential components determine the way that GPTs could learn to generate design concepts. First, knowledge and reasoning for the task is the key component of our framework. It is provided through the dataset used for fine-tuning the GPT-2 model, or in the examples of design concepts for few-shot learning of GPT-3. Secondly, we take in varied input for conditional learning. The input should be customized and consistent with the specific reasoning we want the model to learn, and the output will be the generated design concept description. For instance, for analogy reasoning, the input will be the source and target domains for analogy mapping. Finally, the transformer in the framework can be either a fine-tuned GPT-2 or the pre-trained GPT-3 for few-shot learning.

### 4.2 Concept Evaluation

As discussed in Section 3.4, most NLG evaluation metrics require ground truth references and measure the similarity between the reference and the generated result given the same condition. However, for design concept generation, the aim of the NLG task is not to generate texts that are as close to the reference as possible (as in common NLG tasks like machine translation). On the other hand, we aim to produce novel concepts diverging from existing ones to expand design alternatives. Thus, evaluation metrics will focus on assessing the novelty of the generated concepts.

Traditional approaches toward novelty evaluation require the judgement from human experts (e.g., [101], [102]). However, novelty is subjective to the opinion, intuition, and experience of the evaluators [103, 104]. Human experts can hardly be knowledgeable enough about every existed innovation to give a fully objective novelty assessment. Therefore, data-driven novelty evaluation metrics have been explored in recent years [98, 105, 106]. Especially, the semantic distance or relevancy between terms have been measured based on word embeddings to represent novelty [107, 108]. The word embeddings are vector representations learned from a large corpus of textual data to encode the meaning of the words [109]. This allows the metrics to comprehend the extensive knowledge in the corpus when evaluating novelty. The applicability of semantic distance is also supported empirically by creative cognition research [110].

In this paper, we apply two informative metrics based on semantic distance. One is to use WMD in reverse to measure the diversity between new concepts and existing ones. Prior approaches often aim for minimizing WMD as they want the generated results to be as close to the ground truth as possible. However, we want to do the opposite in our design concept generation task, i.e., assess which model gives the higher WMD score. Higher WMD means the embedding spaces of the generated text and the reference are more distant, thus showing the model can deliver more diverse results. In design tasks, the diversity of concepts could lead to novel solutions, and therefore we use the metrics to represent the model's capability of generating potentially novel concepts. This approach is only applicable when the concepts of the target domain of interest are obtainable so that the generated new concepts can be compared with the reference. The word embedding extraction method for WMD that we use in this paper is the pre-trained Word2Vec model provided by Gensim[2].

Secondly, we propose a within-text evaluation metrics utilizing TechNet [15-16] to measure concept novelty without the need for ground truth references. TechNet can retrieve an adjacency matrix of term-term relevancy from a given text based on the word embeddings. The lower relevancy score means the given terms are more semantically distant in the design context. In the metrics, all term pairs in a generated idea are extracted and the pairs' semantic distances are calculated. Then, the minimum value among them can be used to estimate the degree of novelty regarding the knowledge distance within the idea. The lower minimum score indicates a more novel idea. However, one drawback of this method is that it assumes the generated result being evaluated is concise and only contains critical information about the innovation of the concept. Any extra information (e.g., minor features, style of the product) may interfere with the evaluation by improperly assigning a reduced minimum term-term relevancy score to the concept.

Both evaluation metrics used in this paper are embedding-based because we want to semantically measure the novelty and diversity of the generated concepts. Moreover, both metrics use the bag-of-words representation of the concepts, which means they only measure the content and topic semantically while ignoring the grammar and logic in the texts during evaluation. This allows the evaluation of concepts in different syntaxes or formats.

## 5 Experiments and Results

In this section, we explore the capability of GPT to leverage and transform knowledge and reasoning from design data to new design concepts through three experiments. The settings of experiments are reported in Table 2.

Table 2: Experiment settings

| Experimental Task | Data Source | Input Condition | Transformer Model | Evaluation Method |
| --- | --- | --- | --- | --- |
| Domain Knowledge Synthesis | USPTO patent data (domain-focused) | Target domain | Fine-tuned GPT-2 | TechNet |
|  |  | Target domain | GPT-3 few-shot learning | TechNet |
| Problem-driven reasoning | RedDot award-winning design data | Concept category, problem statement | Fine-tuned GPT-2 | WMD |
| Analogy-driven reasoning |  | Target and source domains | GPT-3 few-shot learning | WMD |

---

[2] https://radimrehurek.com/gensim/

**5.1 Domain Knowledge Synthesis**

Luo et al. [9] conducted the study of design concept generation via domain knowledge synthesis with the help of a stimuli-based tool InnoGPS. Following their work, this experiment focuses on the concept generation of rolling toys synthesizing near-domain and far-domain knowledge. InnoGPS[3] is a knowledge-based design expert system based on all USTPO patents from 1974 to 2020. The system guides the retrieval of near-domain and far-domain stimuli by knowledge distance to aid creative design ideation. The system uses a network of all international patent classes to store, organize and retrieve the world's cumulative technical design knowledge according to statistically estimated knowledge distance between patent classes. In InnoGPS 2.0 [10], the design stimuli are provided at the semantic, document, and field levels simultaneously. While providing data-driven ideation aids, InnoGPS itself does not have generative capability. In this experiment, we employ InnoGPS to gather patent data and assess knowledge distance.

Six domains are picked as references according to their rank order by knowledge proximity to rolling toys: 1) Weapons, 2) Agriculture, 3) Lighting, 4) Drilling & Mining, 5) Grinding & Polishing, and 6) Fuels & Lubricants. Regarding knowledge distance calculation, please refer to [9-10]. The first three domains are relatively near-field (13th, 17th, 28th nearest, respectively) while the other ones are considered far-field (64th, 68th, 100th nearest, respectively). These domains vary a lot regarding the number of patents they hold. To control variables and test for performance, we picked the latest 20,000 patents from each domain to form a dataset. Moreover, the titles of the chosen patents all have more than three words because we want the model to learn to generate ideas that contain rich information. For each domain, we gather the patent titles from InnoGPS and extract a keyword for each title using automatic keyword extraction techniques. Here, we utilize KeyBERT[4] which leverages BERT embeddings to extract keywords and key phrases that best represent a document. Figure 6 shows an example of patent title data prepared for GPTs. The extracted keyword is put in front of the title as the generation condition, together with fixed prompts that indicate the contents. Once the GPT model is prepared after either fine-tuning or few-shot learning, the target domain can be input as the condition to generate new concepts. In this experiment, "rolling toy" is used as the target domain.

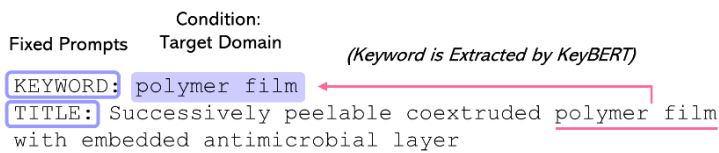

Figure 6: Data preparation example for the domain knowledge synthesis

This study experiments with GPTs for concept generation via domain knowledge synthesis, which was done by human designers in [10]. Both GPT-2 fine-tuning and GPT-3 few-shot learning are tested to compare their performance in learning and leveraging the domain knowledge. For fine-tuning GPT-2, we use the 355M base model, and the model for each domain is fine-tuned for 20,000 steps with a batch size of 1. It takes dozens of minutes to fine-tune each GPT-2 domain experts with a T4 or G100 GPU provided by Google's Colaboratory[5]. Figure 7 shows the fine-tuning loss of each model during the first 10,000 steps. The loss is stabilized afterward. The fine-tuning loss is plotted every 100 steps. The parameters of 'temperature=0.9; top-k=50' are used in the generation to enable a more random generation for more creative ideas. 500 rolling toy design concepts are generated by the virtual domain expert (i.e., the fine-tuned GPT-2 based on each domain's knowledge) and the unique ones are selected for further analysis.

---

[3] http://www.innogps.com/
[4] https://github.com/MaartenGr/KeyBERT
[5] https://research.google.com/colaboratory/

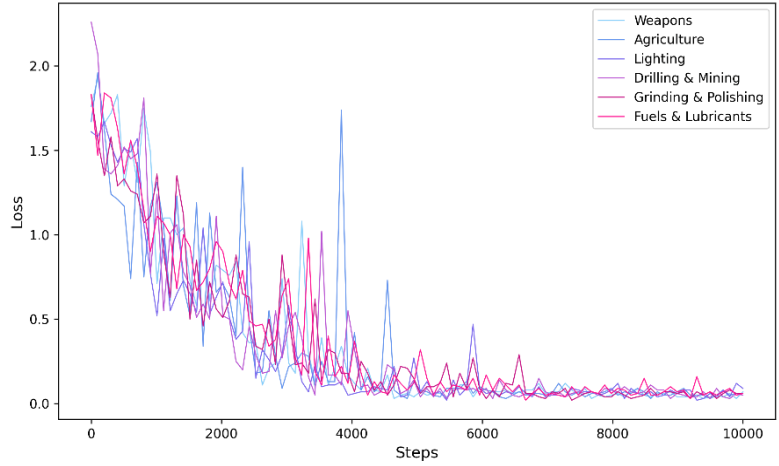

Figure 7: Training loss during fine-tuning the domain-specific GPT-2 models

For GPT-3, 10 samples are randomly selected from the domain data to form the prompts for few-shot learning. Even though 10 sentences contain hardly any knowledge themselves, they can evoke the domain-related knowledge of GPT-3 that was learned during pre-training. Using the GPT-3 base model of 'Curie' with the temperature of 0.9 and top-p of 1, 500 concepts are generated for each domain and the unique ones are selected. Table 3 reports the percentage of unique rolling toy concepts generated by each virtual domain expert based on GPT-2 or GPT-3. Some examples of the generated design concepts are also listed in table 3 to provide an intuition of the what the results are like.

Table 3: Results of rolling toy concept generation

| Domain | GPT-2 | | GPT-3 | |
|---|---|---|---|---|
| | % Unique Ideas | Examples | % Unique Ideas | Examples |
| **Weapons** (13th nearest) | 35.8% (179/500) | • Rolling toy wheeled target.<br>• Rolling toy dart board capable of making turns.<br>• Rolling toy air gun.<br>• Rolling toy dart game.<br>• Rolling toy targeted at delivering a selection of magnetic objects. | 71.8% (359/500) | • Rolling toy having a safety lock.<br>• Rolling toy as firearms accessory.<br>• Rolling toy pistol.<br>• Rolling toy projectile.<br>• Rolling toy with propeller and internal storage. |
| **Agriculture** (17th nearest) | 40.8% (204/500) | • Rolling toy bale wrapper apparatus.<br>• Rolling toy saddle with pressure adjustment.<br>• Rolling toy with liquid container.<br>• Rolling toy splitting system.<br>• Rolling toy harness with attached fabric head. | 67.4% (337/500) | • Rolling toy and a treadmill powered by the rolling toy.<br>• Rolling toy that caters to animal and pet owners.<br>• Rolling toy with multi-directional steering.<br>• Rolling toy having a revolving cylinder comprising interlocking, rotating bands. |
| | | | | • Rolling toy with engaging and disengaging feature. |
| **Lighting** (28th nearest) | 69.6% (348/500) | • Color changing LED roll toy.<br>• Rolling toy and cart with a plurality of removable LED-units.<br>• Lighting device for rolling and adjusting a light source.<br>• Color changing, multi-sided rolling toy.<br>• Programmable multi-color rolling toy. | 84.2% (421/500) | • Rolling toy with illuminated stickers.<br>• Rolling toy for lighting visual display and gaming.<br>• Rolling toy with a lighted pointer feature.<br>• Rolling toys with artificial light sources.<br>• Rolling toy for exhibition model. |
| **Drilling & Mining** (64th nearest) | 69.8% (349/500) | • Rolling toy spindle drive system.<br>• Rolling toy anti-locking system and method.<br>• Rolling toy spinner reel.<br>• Rolling toy deflector and method of use.<br>• Rolling toy milling block and cylinder. | 73.6% (368/500) | • Rolling toy and reel assembly.<br>• Collapsible rolling toy.<br>• Rollover rescue device utilizing a rolling toy.<br>• Rolling toy with three-axis control.<br>• Rolling ball with laser apertures. |
| **Grinding & Polishing** (68th nearest) | 76.4% (382/500) | • Fixture for rolling toy sleeves.<br>• Deep rolling toy arm with interchangeable rolling force.<br>• Nozzle device for the rolling of a rolling toy.<br>• Rubber rolled toy machine with assembly clamping mechanism.<br>• Deep rolling toy arm for a grinding machine. | 77.2% (386/500) | • Toy for rolling on non-slip surfaces<br>• Rolling toy with hidden wheels.<br>• Rolling toy with wheels and axle used for manipulating surfaces.<br>• Rolling toy with a track.<br>• Motorized rolling industrial toy with cam engine. |
| **Fuels & Lubricants** (100th nearest) | 68.6% (343/500) | • Toy diesel fuel production system and rolling toy vehicle.<br>• Toy with rolling bearing and friction mechanism.<br>• Electrical tool and planer for use in a rolling toy.<br>• Lubricant for rolling toy.<br>• Toy diesel fuel and lubricant for use in a rolling toy vehicle. | 84.4% (422/500) | • Apparatus for the rolling of steel sheet to make rolled steel sheet for railway track.<br>• Rolling toy having a pneumatic cylinder.<br>• Flow reactor with a porous rolling toy.<br>• Rolling toy comprising a coating composition.<br>• Method for producing a plastic rolling toy comprising elastomer. |

In general, both GPT-2 and GPT-3 can generate understandable and concise design concepts in natural language. The concepts do take advantage of external knowledge in the design data provided. The results also show that the concepts from the GPT-3 based virtual experts are less repetitive. This can be potentially valuable for the ideation process with both near-field and far-field knowledge sources. For example, given near-field stimuli of weapons, rolling toy designers may quickly come to the idea of "adding a gun or other shooting mechanism to the rolling toy" and then fixate on it. The AI-generated concepts, in addition to that, suggest that the rolling toy can also be designed into a moving shooting target or utilize the idea of a safety lock from the domain of weapons. For far-field stimuli, the virtual experts can also generate ideas as inspiration stimuli when human designers find it challenging to synthesize such distant knowledge. However, as the titles are too simple in syntax, they hardly carry any interesting logical reasoning. Nevertheless, two basic forms of reasoning can be observed from the generated concepts:

- *Use-case-oriented*: Utilize the target in a domain-specific context or application (e.g., "Rolling toy dart board capable of making turns", "Rolling toy that caters to animal and pet owners").

- *Feature-oriented*: Add a domain-specific function or structure to the target (e.g., "Rolling toy with liquid container", "Rolling toy having a safety lock").

However, from the examples shown in Table 3, it is not clear how the different knowledge distances and different GPTs may affect concept generation. According to Luo et al. [9-10, 111] and Srinivasan et al. [112], when ideating with external knowledge, the resulting concepts that synthesize far-field source knowledge could be more novel compared to those utilizing near-field stimuli. It is expected that the concepts generated by GPT-based virtual domain experts will follow a similar pattern, i.e., far-domain virtual experts may generate ideas with higher novelty than near-domain virtual experts. TechNet is used for the evaluation of the generated concepts because similarity-based metrics are not applicable in this experiment. In addition, the AI-generated concept descriptions are concise and typically contain no more than 15 words. This property of concision makes the minimum term-term relevancy score a good indicator of novelty based on the knowledge contained within a text.

Figure 8 reports the evaluation results based on TechNet. In TechNet, as reported in [15], the estimated mean value of the term relevancy score is 0.133 (shown as the horizontal line in Figure 8). Therefore, either near-field or far-field models achieve overall good novelty. Note that this mean value of relevancy was calculated using 108 different pairs of terms. In real contexts, the more relevant pairs will appear more frequently, thus making the mean value even higher. Moreover, for either GPT-2 or GPT-3, concepts generated by the far-field domain expert models achieve generally lower relevancy scores (indicating higher novelty) than those generated by near-field models, thus verifying our expected pattern. However, comparing Figure 8 (a) and (b), this pattern is far less obvious in GPT-3 few-shot learning, probably because it leverages knowledge mostly from its pre-training common sense dataset. In addition, Figure 9 shows that both GPT-2 and GPT-3 based virtual experts tend to generate longer concept descriptions for far-field domains. Longer texts are likely to contain richer details which could also be the source of the lower minimum relevancy because it could increase the probability of the co-occurrence of terms with low relevancy.

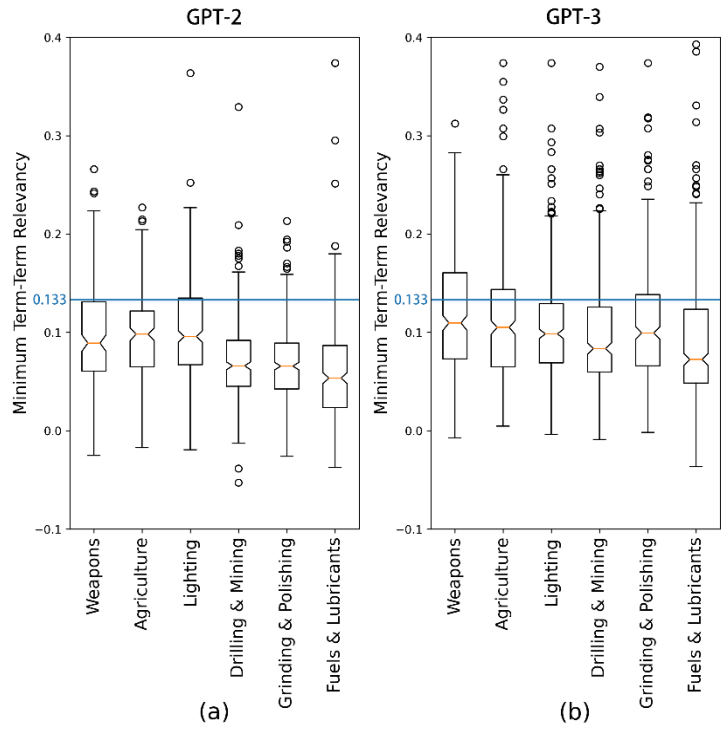

Figure 8: Distribution of the minimum term-term semantic relevancy of concepts generated by GPT-2 (a) and GPT-3 (b)

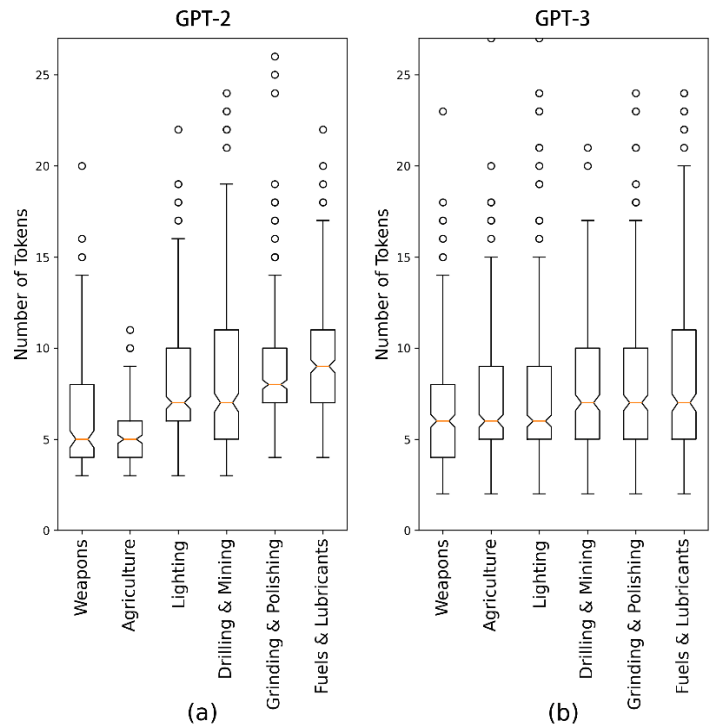

Figure 9: Distribution of the number of tokens of the concepts generated by GPT-2 (a) and GPT-3 (b)

Generative Transformers for Design Concept Generation (A PREPRINT)

Overall, comparing the two GPT models, GPT-2 has the advantage of leveraging a large amount of domain-specific knowledge by fine-tuning. This advantage results in more novel concept generation, especially when synthesizing far-domain knowledge. On the other hand, GPT-3 only requires a few examples to achieve reasonably good performance, and the process does not need further training to update the hyperparameters of the model. Thus, GPT-3 based concept generation can be very easy to use but also not so controllable.

**5.2 Problem-Driven Synthesis**

Given GPT's capability to generate text based on the understanding of the context, we experiment with its application in generating the text of solution ideas for a given problem. Problem-solving in design could be supported by different methods such as analogy and first principle. In this experiment, we test GPT-2's problem-solving performance without constraining its approach, i.e., it is free to use any methods based on what it learned from the data to solve a given design problem. The dataset for model fine-tuning is collected from RedDot's official website[6], including 14,502 product designs from 2011 to 2020 and 1,486 design concepts from 2016 to 2020. We organize the dataset by the categories of designs or concepts. Data preparation includes picking out the text description of each design and adding its category name in front of the description. Figure 10 shows an example of the training data prepared for GPT-2. Part of the concept description is omitted in the figure for space-saving.

```
                    Condition:
Fixed Prompts   Concept Category
                                        Loose Condition: Problem Statement
CATEGORY:  Third Age
DESCRIPTION:  In the narrow and limited space of public
toilet cubicles, old people may accidentally fall when
they squat for too long and stand up too quickly. This
will go unnoticed by people and help cannot be offered
in time when an accident occurs. The Pre-warning
Toilet Door is designed such that an indicator on the
cubicle door will turn yellow and a buzzer will sound
for help if the door is shut for more than 30 minutes.
The indicator shows up green ……
```

Figure 10: Data preparation example for the problem-driven reasoning experiment

The concept category is used as a condition to control for the general domain of concept generation. In addition, the problem statement is introduced as a loose condition, which could express the design problem in varied forms. As shown in the example in Figure 10, the first sentence of the concept description states the phenomenon of an existing use case context. This is followed by the second sentence which proposes potentially problematic or even harmful experiences. The solution idea description follows. This structure is common in the descriptions of problem-driven design and thus ideal for GPT to generate solution idea text as the output in response to a problem text as the input. However, some problem statements come in different forms and do not necessarily contain both the phenomenon and the effect. Moreover, not all award-winning designs are problem-driven and often the description does not begin with a problem description. Our hypothesis is that the model can learn from those varied problem-driven design descriptions (as the example above) in the total dataset to execute problem-solving tasks, while also learning from design descriptions of other structures during training. Therefore, we consider the problem statement as a loose condition in this concept generation experiment and test with different problems. This includes 1) a known problem to a known solution, 2) a new unsolved problem, and 3) an open problem that includes only the target product. Among them, 3) is not a complete problem statement and relies on the model to determine novel needs based on its knowledge.

Using the entire RedDot dataset, a pre-trained GPT-2 of 355M parameters is fine-tuned for 28,000 steps with a batch size of 1. It takes couple of hours to fine-tune the GPT-2 model with a T4 or G100

---

[6] https://www.red-dot.org/

GPU provided by Google's Colaboratory. Figure 11 reports loss over the fine-tuning steps, added with a simple moving average (SMA) plot over 2000 steps.

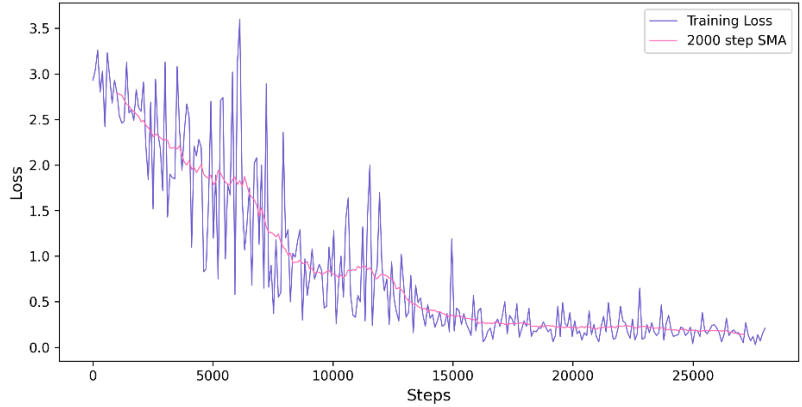

Figure 11: Training loss during fine-tuning GPT-2 using RedDot award-winning data

The problem used for the experiment is based on an IF award-winning design concept in 2021 which proposed a wireless ECG device to improve the medical testing experience for children[7]. After minor simplification, the problem statement of "current electrocardiograph testing involves a large number of wires, making the process time-consuming for doctors and stressful for children" is used as the input of a known problem. This concept is not included in the fine-tuning dataset and the pre-training dataset of GPT-2 does not include any data after 2019. Thus, even if we do not know the pre-training dataset composition, it is certain that the fine-tuned model has no knowledge of the solution to the problem. Based on the same target product, the new problem is defined as: "Current electrocardiograph testing may cause tissue breakdown or skin irritation if the electrode patches are left on too long" which is from a health advice from a medical organization[8]. In addition, the open problem input is simply as "Current electrocardiograph testing" which only define the target and leaves the problem space open for the transformer to complete. Certain starting tokens such as "current" or "existing" could evoke the fine-tuned model to complete the problem statement with its own knowledge before generating any solutions. This could allow novel needs of the target product to be explored. To test performances, 500 concepts are generated for each problem prompt using the parameter of 'temperature=0.9; top-k=50'. The input condition of the concept category in this experiment is fixed as "Life Science". Table 4 reports three examples of the generated concepts for each problem statement.

Table 4: Examples of problem-driven ECG device concept generation

| Input Problem | Examples of Generated Concepts |
|---|---|
| Original Problem | 1# Current electrocardiograph testing involves a large number of wires, making the process time-consuming for doctors and stressful for children. The Patchboard electrocardiograph features an innovative retractable electrode cardiopulae that retracts into the paper cartridge during testing. With the help of an adaptable patch board, the doctor can now safely excorlate and patch the data up onto the child's bed. The cartridge is retractable and fixed in place at the base of the electrocardiograph, making it convenient to leave the cartridge lying around. The design of the electrocardiograph patchboard is adaptable to every child. It is sized to fit every child's height and is made from a variety of materials to be used in craft and manufacturing industries. |
| | 2# Current electrocardiograph testing involves a large number of wires, making the process time-consuming for doctors and stressful for children. Therefore, the EM 20– package is designed to replace traditional testing by a wireless device that affords an optimal fit for children's clothing. To wear, adjust |

---

[7] https://ifdesign.com/en/winner-ranking/project/ema/324944
[8] https://www.hopkinsmedicine.org/health/treatment-tests-and-therapies/electrocardiogram

| | |
|---|---|
| | the button to the age of the child and attach the other device to the chest. Then, insert the child's birthday cake into the wireless device. The birthday cake will remain attached to the device throughout the entire testing process, allowing doctors to continue the testing procedure even after the cake has been removed from the device. This device is also designed to fit children's clothing in their regular bra size. |
| | **3#** Current electrocardiograph testing involves a large number of wires, making the process time-consuming for doctors and stressful for children. Future Trig Therapy device could be used to help reduce the frequency of testing and thereby increase the doctor's comfort level with patients. This future device could be worn like a fashion accessory or a 'glimmer' – a magnetic coil. Instead of testing electrical impulses in a baby's arm, it stimulates the body through electromagnetic waves. In its younger stages, the electromagnetic waves are transmitted to the muscle tissue and the signals are converted to electrical impulses. As the doctor continues to work in the arm, the electrocardiogram can be better monitored. |
| New Problem | **4#** Current electrocardiograph testing may cause tissue breakdown or skin irritation if the electrode patches are left on too long. Long-term use of the device can result in cognitive impairment, resulting in discontinuation of the generator therapy. ARTISTICIZE patches are designed to address these issues. ARTISTICIZE is a new cartridge type of electrocardiograph that includes an embedded patch board in which long-term use of the device can continue without cognitive or learning impairment. This allows the body to continue producing clotting factor, thereby meeting the demands for its primary organ. The ARTISTICIZE cartridge board is engineered to provide an optimal electrode distribution for the long-term use of electrocardiogram. The long-term use of the device results in high levels of treatment, which are well-thought-out in terms of both symptom severity and dosimeter compliance. The ARTISTICIZE cartridge is well-thought-out because it incorporates high-quality materials that provide fine-motorable soft tissue augmentation. |
| | **5#** Current electrocardiograph testing may cause tissue breakdown or skin irritation if the electrode patches are left on too long. To address this problem, the Five Eyes Patch devices use a unique algorithm to monitor the electrical activity in the muscle layers and automatically adjust patch length and electrodes applied. This not only makes for more precise patch placement, but it also reduces the risk of scarring and infection. The product can be applied topically, or it can be attached to the wrist via a strap. |
| | **6#** Current electrocardiograph testing may cause tissue breakdown or skin irritation if the electrode patches are left on too long. To address this problem, Patch T130 shows a duration of 510 micros (micros) that can be applied along the along electrode patch length. The thickness of the patch can be adjusted to the size of muscle sheaths to protect both the hypertrophied patient and the accessory device. The adhesive patches are made of medical-grade, skin-friendly tissue that stays on the patient patch and is removed when the device is attached to the skin. |
| Open Problem | **7#** Current electrocardiograph testing equipment is bulky, not aesthetically pleasing and generally not ergonomically optimised. The new NEXT generation of electrocardiograph without electrodes will improve the user experience in a range of areas of daily use, and is also significantly less bulky than current electrocardiograph test systems. New features include a 5G wireless biorhythm sensor and a colour design that allows the biorhythm sensor to be inserted into a variety of fabrics. This makes it easier to insert, and thus more convenient for patients who are transgender or genderfluid. Additionally, a biochip on the back can be added, which signifies that the device is in good working condition. |
| | **8#** Current electrocardiograph testing devices do not allow for long-time recording electrocardiograms without taking multiple body measurements. This time-consuming approach is detrimental to the fitness of recovering athletes who require daily monitoring to be able to play a full part in their rehab and recovery. Testing at Home Is an Option That Can Recorder Next Cycle Fitness will allow you to play a recording session during which you can recharge your device. The product also sends feedback on the exercise completed and calorie consumption is calculated. In addition, it can be connected to a wearable coaching platform on Wear OS so that a personalised coaching tool is available for consumers. |
| | **9#** Current electrocardiograph testing devices do not allow for long-term monitoring of the heart function. This device can send data via Bluetooth link to the owner's smartphone. The testing device is lightweight, small, comfortable to wear and also reduces the agony of wasted testing days. To start using, the owner simply inserts his or her phone into the device andottoads for the appropriate heart function test. Depending on the test result, the owner may be awarded points for a successful or unsuccessful test. |

Intuitively, the model is capable of generating novel and useful concepts for all three problem settings. For the original problem, solutions are proposed to innovate the interaction or materials of the ECG device to be more children friendly. Some of them (e.g., 2#) even mention using wireless technology just like the original award-winning concept even though the model has no previous knowledge about such a solution. For the newly defined problem, the model can either add a negative effect to the given phenomenon before generating new concepts (e.g., 4#) or directly give a solution (e.g., 5#). This variation of details in the problem statement is also observed in the generation based on an open-ended problem. Another noteworthy observation is that it may automatically search for use cases for a solution, e.g., concept 8# mentions that recovering athletes who require daily monitoring could use such an ECG device.

However, one potential drawback is that the machine-generated or completed problem statement is not guaranteed to be objectively correct, and the designers need to check if the problem is valid before moving forward. For example, the problem in 8# mentioned "multiple body measurements", which makes little sense for ECG testing. Nevertheless, the partially correct problem in 8# is still valid and the generated solution could still be useful. Furthermore, as the text lengthens, they are likely to cover other features that are unrelated to the given problem. This is not surprising as the design descriptions in the fine-tuning dataset often provide comprehensive elaboration of multiple aspects and the model learned this well. This is also the major reason that the evaluation metric using TechNet is not applicable for this experiment.

In this experiment, WMD is used for evaluating the diversity of the generated results. We set two ground-truth references for the evaluation, one is the original IF concept as an existing novel design, and the other is an introduction of ECG by Wiki as a common-sense reference. Figure 12 reports the evaluation result of all the 1,500 generated concepts. The results show the concepts derived from all problem settings achieve equally good novelty compared to our common-sense cognition of ECG testing. However, when compared to the original IF concept, the AI-generated design concepts using the original problem setting achieve much lower novelty. This is because the same problem is more likely to lead to similar solutions even for GPT-2. In addition, the concepts with open-ended problems also have relatively lower WMD scores compared to either reference.

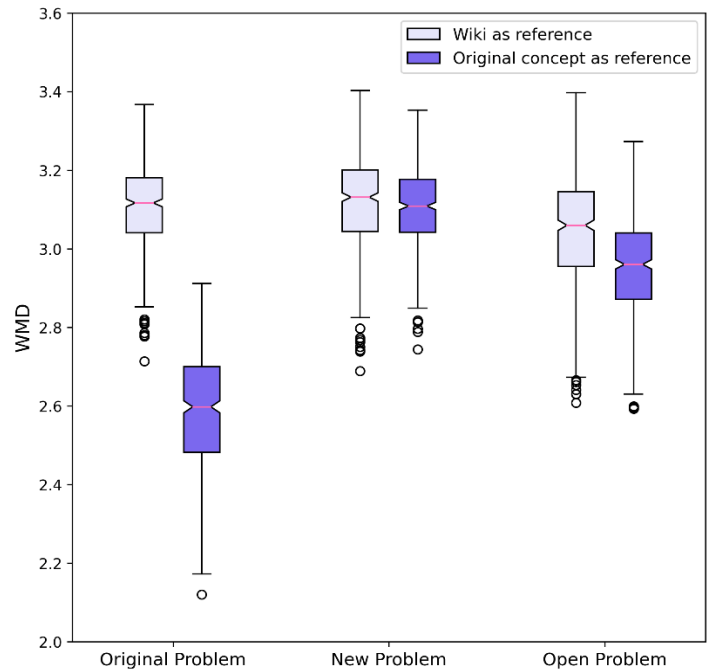

Figure 12: Distribution of the WMD scores of the generated problem-driven concepts comparing to different references

### 5.3 Analogy-Driven Synthesis

Design-by-analogy is the projection of existing reference in a source domain to address a comparable challenge in the target domain [16, 42, 113]. It is usually considered a problem-solving approach. However, when a problem is not specified, analogy reasoning can also lead to open-ended design concept generation. This experiment is to test the analogy-driven reasoning of the model for concept generation. The experiment is guided by TechNet to explore near-domain and far-domain analogy by the semantic distance between the source and target term. This is not only applicable for the context when a designer aims to establish clear analogy mapping across given domains, but also for exploring unfamiliar or unexpected domains as analogies.

As there are insufficient design-by-analogy examples to fine-tune a GPT-2, this experiment selects from RedDot five analogy-driven reasoning examples for GPT-3's few-shot learning. Table 5 shows the source and target domains in each of the five examples for learning. Before the text description of each sample, a structured sentence specifying the source and target domains (e.g., "Applying accordion to computer mouse") is inserted so that GPT-3 may learn to develop ideas based on the input domains. An example of the prepared sample for few-shot learning is shown in Figure 13. When generating new concepts, we simply need to update the tokens that specify the source and target domains in the condition.

Table 5: Analogy-driven reasoning examples used as prompts for GPT-3

| **Example** | **Source Domain** | **Target Domain** |
|---|---|---|
| *AMBI-*[9] | Accordion | Computer Mouse |
| *Build Fender*[10] | Cells | Building |
| *Sole*[11] | Standing desk | Automobile |
| *FOLD (Light wheelchair)*[12] | Folding chair | Wheelchair |
| *CableTread*[13] | Circuit board | Desk |

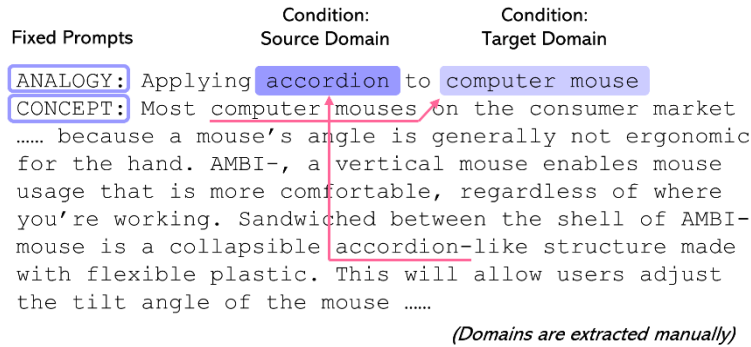

Figure 13: Data preparation example for the analogy-driven reasoning experiment

In this experiment, we fix the target domain as "drone", and select source domains according to their semantic relevancy to the target as well as their own categories. First, one of the most near-field terms

---

[9] https://www.red-dot.org/project/ambi-48504
[10] https://www.red-dot.org/project/build-fender-27044
[11] https://www.red-dot.org/project/sole-26525
[12] https://www.red-dot.org/project/fold-light-wheelchair-26521
[13] https://www.red-dot.org/project/cabletread-46563

"airplane" (relevancy score: 0.421) is selected as the comparison set of the experiment. Next, "bird" and "insect" are picked to test for bio-inspired design (BID). These biological terms have surprisingly high relevancy scores to "drone", which are 0.313 and 0.234 respectively, showing that the BID of drones using birds or insects has already been implemented in engineering design. The last group contains the culture-related terms "lantern" and "poetry" which only have the relevancy score of 0.142 and 0.106. Using the GPT-3 base model of 'Davinci' with the temperature of 0.9 and top-p of 1, we generated 100 concepts based on the analogy of each source domain, three of which are reported in Table 6.

Table 6: Examples of analogy-driven drone concept generation

| Source Domain | Examples of Generated Concepts |
| --- | --- |
| Airplane (0.421) | **1#** The UAV is a drone inspired by modern day aircraft. It is made entirely out of carbon fibres reinforced with glass fibres. This design makes the drone lighter and more aerodynamic in flight. The drone is battery powered and is equipped with a camera that transmits back images to its remote controller. |
| | **2#** Inspired by the interior design of modern airplanes, PLANE is a drone that aims to eliminate the social stigma that drones are 'murder machines'. The green colour and minimalistic style shows a humanised touch, bringing a user-friendly experience to the flying drones. To teach drones the value of an individual's safety and privacy, PLANE is promoted as a drone with 'eyes' – sensors that protect its surroundings. In addition, the heart-shaped, exposed camera on its belly shows the drone's ability to document and deliver information to users. |
| | **3#** Inspired by the efficiency of airplane cargo carriers, the Rotary UAV was designed to transport heavier loads through urban areas while being unobtrusive to citizens. The cylindrical structure of the Rotary UAV provides greater stability and transportation capacity. Funded by the Singapore-MIT Alliance for Research and Technology, the Rotary UAV's rotary engine also provides its ability to drive in any direction, thus eliminating the need for a runway. It is propelled by a motor spinning the rotor of the plane (vertically oriented). This technology would make it possible to transport heavier items or cargo around a neighbourhood without a helipad or runway. |
| Bird (0.313) | **4#** The body structure of aircraft must be redesigned to accommodate a greater number of passengers and cargo. SPEA is modelled after a bird, specifically the blue-footed booby to achieve speed and stability in flight. The reconfiguration of the wings and tail has helped achieve this goal. APASI was chosen as it allows more space under the wings in comparison to other designs which helps reduce drag force. |
| | **5#** DSDK is inspired by the way birds flap their wings in loop patterns before taking off. The 'Bird-like drone' is able to maintain stability and direction during flight whilst allowing quick movement, fluidity, and coordination. Using this model, the drone is able to travel long distances and still be able to maintain or adjust its position. Self-deploying drones that incorporate functionalities like this are useful for many applications, for example delivery. The self-deploying nature of the drone allows it to be dropped from a safe distance and travel at a steady pace without much piloting needed. |
| | **6#** DUCOPTER is inspired by birds. It integrates two vertical takeoff and landing capabilities. The launch and landing process are similar to how birds take off and land during flight. The first is the familiar takeoff – propellers are turned on, and the wings provide lift. After reaching a certain height, the wings fold down, allowing the aircraft to move forward like a bird gliding in the air. This innovation will allow the aircraft to land safely when the battery dies. |
| Insect (0.234) | **7#** Inspired by the efficiency of the ant colony, a multi-user drone was conceived. The drone can be controlled by a single user for one task and also by several users for another, to perform tasks together. Additionally, the drone contains an "ant defense mechanism", which is inspired by ants' capacity to self-heal. The mechanism enables other drones that are functional to return home while delivering spare parts to repair the broken down drones. |
| | **8#** Inspired by the hard and heavy exoskeletons of beetles and wasps, PROTECTOR is a drone designed for military surveillance operations. The thick outer casing is made of either ceramic or polycarbonate and armour plate protection, an essential extra protection which will help protect the drone and increase its chance of survival in hostile environments. The drone is powered by two |
| | batteries with a capacity of 200Wh, which can fly for 12 hours and run the surveillance equipment for up to 20 hours. |
| | **9#** Insects have the ability to communicate and navigate through vibration. Inspired by this, Bugs² is a drone that produces on-board vibration to communicate with other Bugs² units and effects. Lower frequency vibration will be used to communicate with other Bugs², while higher frequency vibration will be used to communicate with other aircraft nearby. The system will also produce a unique vibration frequency when the drone has detected obstacles nearby, sending this signal to other drones nearby. This will create a swarm that flies together, avoiding any collision. |
| Lantern (0.142) | **10#** Inspired by traditional Chinese paper lantern, which is made of just three steps and a hexagonal wire frame. Drone is a drone which can be assembled simply with a hexagonal wire frame using extremely flexible, heat-resistant, and dual-ended tape. The frame is then filled with a flexible, heat-resistant material. The filling can be varied to make different shapes of lanterns, such as one candle-like and another resembling a sun. |
| | **11#** Night flying using drones is highly restricted due to the lack of visibility of the surroundings and its low altitude. LampTread is a drone concept that can transform into a lantern. It is designed with a hook that attaches to a drone to be flown at night. By tapping a button, the hook allows the drone to be hung from lamp posts that are on both ends of a road intersection. This makes it easy for users to identify locations at night by just looking up. |
| | **12#** The lantern for drone is inspired by vernacular fishing trap lanterns designed to attract fish to shallow water to be easily harvested. The concept transforms the conventional UFO-looking drone into a translucent and glow-in-the-dark lantern. The lantern not only attracts fish, but it also attracts people in the area for them to be aware of the existence of the drone, hence the drone does not need extra lights to create an illuminated light environment. It also means that unwanted attention brought by the drone is mitigated. |
| Poetry (0.106) | **13#** Why would we use rigid laws and encyclopedias to define the world? Epic# is a multi-functional swarming drone inspired by the idea of poetry. To create a unified atmosphere or shared experiences, the team behind this concept hopes to develop an AI system for drones. The AI will generate different types of poetic stories based on real-time data collected from sensors and cameras – such as weather conditions and citizens' emotions. On a conceptual level, Epic# can be used to explore new forms of language while communicating despite different cultural contexts. |
| | **14#** Inspired by the Japanese art of Haikai, a light drone is imagined to use during outdoor gatherings. When people feel that their conversations have been exhausted, a set of light drones will rise from the ground and emit cute words with lights that quickly disappear – giving new life to the conversation. The drones are designed to have a simplistic structure and use a single circuit board and propellers. Its polycarbonate shell is soft and tough enough to survive even after hitting a target. |
| | **15#** Inspired by poetry, the project is set to redesign the popular Chinese consumer drone, HeChat, by introducing a poetry element with the aim of broadening the user's range of expression. Poetry is a form of art that revolves around the expression of thoughts and feelings through words. In this concept, a poetic expression module is introduced. Instead of programming the robot to follow a set sequence of motions, this module creates a state-machines based movement that relies on an improvisation mechanism. This improves the user experience by providing an infinite number of movement sequences the user can manipulate. In the end, this makes the drone a fun companion that can improvise by the user's movements. |

As expected, the domain of airplanes does not provide many novel insights for drones because the mapping between the two domains already exists to a great extent. For biological analogies, GPT-3 can not only retrieve a specific creature from the given domain for inspiration but also learn behavioral or structural features from the creatures. For example, 7# proposes to innovate the control system of the swarm of drones based on the group behavior of ants, 8# takes inspiration from the exoskeleton of beetles to design tough military drones. For the culture-related terms as analogies, the model is also capable of generating interestingly and surprisingly novel concepts from varied angles. When using the term lantern as the analogy, 10# retrieves the hexagonal wire frame structure of the Chinese traditional lantern to design an easy-to-assemble drone. 11# proposes a lighting drone that can either fly or hang at fixed positions to light the surrounding. Interestingly, the idea of 11# is very similar to an IF award-winning drone concept[14]. Finally, poetry could be a hard-to-imagine analogy for human designers because of its intangible and artistic nature. 13# suggests using swarm drones to collect real-time sensor and camera data to generate poetic stories, 14# also proposes an interesting design of drones for social activity.

Overall, the results show promising performance of GPT-3 to implement analogy-driven reasoning. However, the generated concepts face a similar issue as the ones from the previous experiment, which is that they are not concise enough and contain trivial information. Thus, for the quantitative evaluation of the method, WMD is utilized. In this experiment, we form a common-sense reference from Wiki's introduction to the unmanned aerial vehicle (UAV), and an existing novel concept reference consisting of four RedDot award-winning drone concepts in recent years. The evaluation results are shown in Figure 14. The four award-winning concepts to form the reference are: Bladeless Drone[15], Xenon Adventure Drone[16], Racer[17], and Rotate and Fly[18].

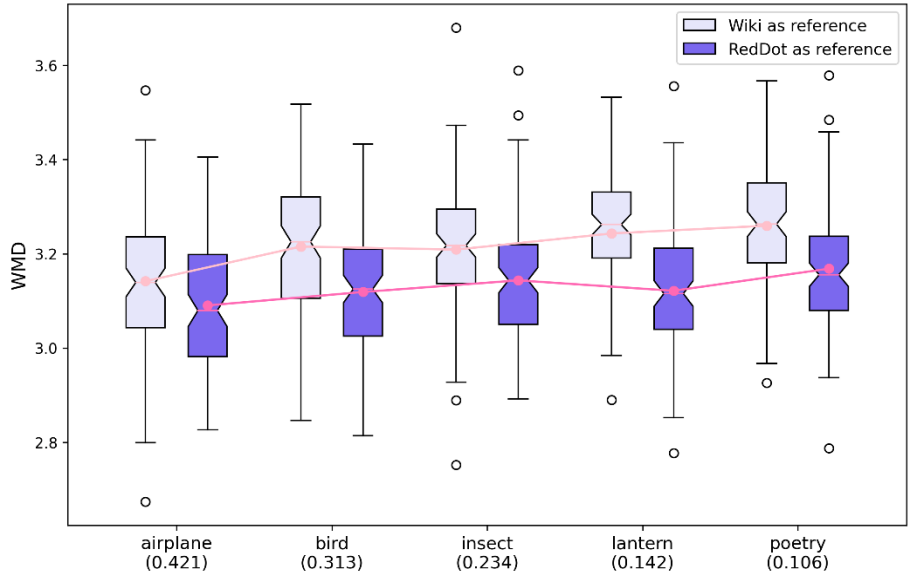

Figure 14: Distribution of the WMD scores of the generated analogy-driven concepts comparing to different references

Evaluation results show that the generated concepts are generally more novel compared to the Wiki knowledge than comparing to RedDot concepts, which is not surprising because concepts on RedDot already represent a high-level novelty of design (calculating the WMD score of the two references gets the result of 3.159). Our generated concepts are shown to achieve better novelty than existing novel concepts. In addition, with the source and target domains getting semantically distant, the generated results also show increasing novelty. However, this trend is observable but not obvious, similar to the results of our GPT-3 experiment in Section 5.1. This is most likely due to the fact that GPT-3 mainly leverages common-sense knowledge during few-shot learning and little technical knowledge is introduced to enhance diversity further.

---

[14] https://ifdesign.com/en/winner-ranking/project/illuminated-drone/333309
[15] https://www.red-dot.org/project/bladeless-drone-26825
[16] https://www.red-dot.org/project/xenon-adventure-drone-27016
[17] https://www.red-dot.org/project/racer-40463
[18] https://www.red-dot.org/project/rotate-and-fly-54390

# 6  Discussion

To the best of our knowledge, this paper is the first to explore and experiment with the uses of generative transformers (i.e., fine-tuned GPT-2 or GPT-3 few-shot learning with design data) for generating novel and useful design concepts in the early design stage. Our first experiment (focusing on domain knowledge synthesis for concept generation) shows that the generative transformers based on GPT-2 have greater capability to leverage domain-specific expert knowledge learned during fine-tuning, especially when synthesizing far-domain knowledge, than GPT-3. This indicates that GPT-2 can be an ideal option when the design task is highly domain-specific, and at the same time, a large amount of domain knowledge is available.

In the second experiment (focusing on problem-driven synthesis for concept generation), GPT-2 can also learn specific design reasoning even if not all samples in the fine-tuning dataset contain such type of reasoning. In addition, novel needs that were previously unknown to the designers can be generated when the problem statement is not strictly defined. This makes the interaction with computer more flexible when exploring design opportunities using generative transformers.

On the other hand, the experiment (focusing on analogy-driven synthesis for concept generation) reveals the performance of transformers based on GPT-3 to leverage a much larger variety of common-sense knowledge and reasoning than those of GPT-2. GPT-3 only requires a few examples of the expected reasoning. The exploration of initial design opportunities often starts with a wide-open design space and technical expert knowledge is seldomly required until later process. GPT-3 can be a valuable and easy-to-use tool in this context.

In sum, the readily available pre-trained language model, the collection and customization of the dataset for fine-tuning or few-shot learning, and the development and selection of evaluation metrics are all task-specific. This makes task definition and training data customization essential steps when it comes to utilizing the PLM for design concept generation. Based on the experiments conducted in this paper, we propose three guidelines for defining the concept generation task and preparing the training dataset:

1) Define the conceptual design task as a combination of conditions and completion. The conditions can be any constraints or requirements defined during the task clarification stage of design process. However, do note that GPTs can hardly comprehend every detailed constraint and requirement, so the conditions should be selective and explorative.

2) Collect and curate a dataset that contains the defined conditions and completion. Extract the conditions manually or with automatic NLP techniques like keyword extraction or named-entity recognition. If a condition is not available in every sample, try to utilize it as a loose condition.

3) When customizing the dataset, always put the conditions at the beginning of the text. Because GPTs are auto-regressive models that predict the next token based on all tokens before.

Despite the observable outstanding performance in concept generation with fine-tuning and few-shot learning, exploiting generative transformers from scratch can be very challenging for engineering design applications. This is because training a generative transformer from scratch requires both a massive amount of data and the super computing power. The lack of openly available engineering design data [31] and powerful but affordable GPU servers can be a major limitation for design researchers to exploiting transformers for meaningful engineering applications.

Furthermore, transformer is a rapid developing technology in the AI community. Although it was originally proposed for NLP tasks [74], experimental works have developed transformers with and for computer vision (CV) in varied ways. For examples, OpenAI published CLIP for zero-shot downstream CV tasks [114] and DALL-E for zero-shot text-to-image generation [115]. Vision transformer (ViT) is also shown to outperform the convolutional neural networks (CNN) architecture in many tasks [116]. A more recent work even achieved high performance image completion on random patches [117]. Generative transformers for CV could open great opportunities for engineering design by not only improving existing deep generative approaches, but also exploring novel human-AI collaborative design methods. Imagine the generated concepts in this paper can be directly transformed into design images through text-to-image generation.

## 7 Limitation and Future Work

Although our work has shown the great performance of generative transformers in design concept generation, there are limitations that need further attentions. We organize four major limitations in this section, namely: interpretability, generalizability, extendibility, and lack of baselines.

*Interpretability*. One major limitation is that the generated concepts in each experiment are not guaranteed to be valid or of good quality. Developing automatic metrics to evaluate and filter the concepts can be very challenging because the quantitative scale of the validity of verbally expressed concepts has yet to be defined. A concept that makes perfect sense is often not insightful, meanwhile ambiguous concepts could carry critical inspirations for designers. Simply filtering out the concepts that could contain key insights just because they are not completely interpretable is not supposed to be the way. Future work will focus on developing an interpretability scale that measures "to what extent the generated concept makes sense". Then we can explore the correlation between interpretability and the actual novelty and usefulness of the generated concepts.

*Generalizability*. To facilitate the design concept generation in real-world design practices, different tasks and the associated knowledge are needed to be generalized. A practical design concept generation process could begin with either a problem from the user's perspective (as in section 5.2), or a technology to be explored with extended opportunities (as in section 5.1 and 5.3). It is necessary for a fully automated AI assistant to comprehend different strategies and learn to pick up the applicable ones for customized input. Given a dataset with adequate size and variety of design knowledge and reasoning, it could be possible to train a model that generalize many tasks. This limitation will be explored in future works.

*Extendibility*. Although the proposed framework in Figure 5 is not limited to the three case studies in this paper, it could be challenging to extend the method to different tasks. This is because the lack of open and high-quality engineering design dataset [31] could limit the method to learn and leverage extended data. However, by customizing the available data, further design reasonings (e.g., reverse thinking) could be learned from the introduced data sources (i.e., patents, award-winning design) and thus extend the applicability of the proposed framework.

*Lack of baselines*. This work is an explorative study and the first to apply AI to comprehend design knowledge and reasoning and generate concepts in natural language. Thus, the presented research lacks existing methods as the comparison set to evaluate performance. Specifically, the metrics to quantitively and automatically evaluate concept novelty and diversity in this paper are specifically tailored to our concept generation tasks. The purpose of these metrics is to compare the performance between different experiment attempts (e.g., could far-field domains lead to more novel concepts compare to near-field ones). Therefore, baseline verification is absent, i.e., how high the WMD score or how low the TechNet relevancy score represents good novelty? There also exist other metrics to automatically evaluate novelty and additional dimensions of new design concepts (e.g., [107]). More extensive experiments on varied subjects and alternative evaluation metrics with human evaluation are needed to discover more insightful patterns or guidelines for using generative transformers in design concept generation.

This work addresses the recent call for advancing NLP and NLG for-and-in design research [65] and especially utilizing the generative transformers in generative design [11]. In the time that transformer-based foundation models [118] start to reshape both academia and industry, we believe it is timely and crucial to explore the opportunities that it offers to creative design. By showing evidence of great potentials, the authors hope this paper to be path-breaking and able to inspire more studies into this area.

Transformer technologies and the new pre-trained transformers (including the vision transformers) are rapidly evolving. We need to continually explore, experiment, and adapt the new ones for applications in the creative design process toward the future. Another important avenue for future research is the design of the process of interactions between human designers and generative transformers. In sum, we hope readers view this paper as an invitation for more research and development of generative transformers in design research and practice.

Generative Transformers for Design Concept Generation (A PREPRINT)[64] Kingma, D. P., & Welling, M., 2013, "Auto-encoding variational bayes," arXiv preprint arXiv:1312.6114. DOI: 10.48550/arXiv.1312.6114

[65] Siddharth, L., Blessing, L., & Luo, J., 2022, "Natural language processing in-and-for design research," Design Science, 8, E21. Doi:10.1017/dsj.2022.16

[66] Gatt, A., & Krahmer, E., 2018, "Survey of the state of the art in natural language generation: Core tasks, applications and evaluation," Journal of Artificial Intelligence Research, 61, pp. 65-170. DOI: 10.1613/jair.5477

[67] Lopez, A., 2008, "Statistical machine translation," ACM Computing Surveys (CSUR), 40(3), pp. 1-49. DOI: 10.1145/1380584.1380586

[68] Ozsoy, M. G., Alpaslan, F. N., & Cicekli, I., 2011, "Text summarization using latent semantic analysis," Journal of Information Science, 37(4), pp. 405-417. DOI: 10.1177%2F0165551511408848

[69] Li, Z., Jiang, X., Shang, L. and Li, H., 2018, "Paraphrase Generation with Deep Reinforcement Learning," Proceedings of the 2018 Conference on Empirical Methods in Natural Language Processing, pp. 3865-3878. DOI: 10.18653/v1/D18-1421

[70] Leppänen, L., Munezero, M., Granroth-Wilding, M., & Toivonen, H., 2017, "Data-driven news generation for automated journalism," Proceedings of the 10th International Conference on Natural Language Generation, pp. 188-197. DOI: 10.18653/v1/W17-3528

[71] Wanner, L., Bosch, H., Bouayad‐Agha, N., Casamayor, G., Ertl, T., Hilbring, D., ... & Vrochidis, S., 2015, "Getting the environmental information across: from the Web to the user," Expert Systems, 32(3), pp. 405-432. DOI: 10.1111/exsy.12100

[72] Perera, R., & Nand, P., 2017, "Recent advances in natural language generation: A survey and classification of the empirical literature," Computing and Informatics, 36(1), 1-32.

[73] Duan, J., Zhao, H., Zhou, Q., Qiu, M., & Liu, M., 2020, "A Study of Pre-trained Language Models in Natural Language Processing," 2020 IEEE International Conference on Smart Cloud (SmartCloud), pp. 116-121. DOI: 10.1109/SmartCloud49737.2020.00030

[74] Vaswani, A., Shazeer, N., Parmar, N., Uszkoreit, J., Jones, L., Gomez, A. N., ... & Polosukhin, I., 2017, "Attention is all you need," Advances in Neural Information Processing Systems 30 (NeurIPS 2017).

[75] Topal, M. O., Bas, A., & van Heerden, I., 2021, "Exploring transformers in natural language generation: GPT, BERT, and XLNET," International Conference on Interdisciplinary Applications of AI (ICIDAAI)

[76] Pascanu, R., Mikolov, T., & Bengio, Y., 2013, "On the difficulty of training recurrent neural networks," International conference on machine learning, pp. 1310-1318, PMLR.

[77] Brown, T. B., Mann, B., Ryder, N., Subbiah, M., Kaplan, J., et al., 2020, "Language models are few-shot learners," Advances in Neural Information Processing Systems 33 (NeurIPS 2020).

[78] Radford, A., Wu, J., Child, R., Luan, D., Amodei, D., & Sutskever, I., 2019, "Language models are unsupervised multitask learners," OpenAI blog, 1(8), 9.

[79] Peng, B., Zhu, C., Li, C., Li, X., Li, J., Zeng, M., & Gao, J., 2020, "Few-shot Natural Language Generation for Task-Oriented Dialog," Findings of the Association for Computational Linguistics: EMNLP 2020, pp. 172-182. DOI: 10.18653/v1/2020.findings-emnlp.17

[80] Amin-Nejad, A., Ive, J., & Velupillai, S., 2020, "Exploring transformer text generation for medical dataset augmentation," Proceedings of the 12th Language Resources and Evaluation Conference, pp. 4699-4708.
30